\documentclass[letterpaper, 10 pt, conference]{ieeeconf}  % Comment this line out if you need a4paper

\IEEEoverridecommandlockouts                              % This command is only needed if 
\usepackage{graphicx} % for pdf, bitmapped graphics files
\usepackage[utf8]{inputenc}
\usepackage{amsmath,amsfonts,booktabs,cite,bm} % general useful stuff
\usepackage{gensymb} % for \degree
\usepackage{siunitx,textcomp} % for the degree symbol
\usepackage[hidelinks]{hyperref} % to have hyperlinks and for the \ref macros
\usepackage{cleveref} % for easier referencing
\usepackage{subcaption}
\usepackage{tabularx}
\usepackage{pbox}
\let\labelindent\relax
\usepackage[inline]{enumitem} % inline enumerate
\usepackage[nolist]{acronym} % acronyms by the \ac{label} command
\usepackage{multirow}
\usepackage[ruled,vlined]{algorithm2e}

\usepackage[font=small]{caption}

\def\secref#1{Sec.~\ref{#1}}
\def\figref#1{Fig.~\ref{#1}}
\def\tabref#1{Tab.~\ref{#1}}
\def\eqref#1{Eq.~(\ref{#1})}
\def\algref#1{Alg.~\ref{#1}}

\newcommand{\matr}[1]{\mathbf{#1}}

\newcommand\etal{~\emph{et al. }}
\newlength{\twosubht}
\newsavebox{\twosubbox}

\crefname{algocf}{alg.}{algs.}
\Crefname{algocf}{Algorithm}{Algorithms}

\title{\LARGE \bf
Combining Local and Global Viewpoint Planning for Fruit Coverage
}

\author{Tobias Zaenker \and  Chris Lehnert \and  Chris McCool \and Maren Bennewitz% <-this % stops a space
\thanks{This work was funded by the Deutsche Forschungsgemeinschaft
  (DFG, German Research Foundation) under Germany’s Excellence
  Strategy – EXC 2070 – 390732324. T. Zaenker, C. McCool, and M. Bennewitz are with the University
  of Bonn, Germany. C. Lehnert is with the Queensland University of Technology (QUT), Brisbane, Australia.
  \newline 978-1-6654-1213-1/21/\$31.00 \textcopyright 2021 IEEE}%
}

\begin{document}

\maketitle
\thispagestyle{empty} 
\pagestyle{empty}

\begin{abstract} 
	
  Obtaining 3D sensor data of complete plants or plant parts (e.g., the crop or fruit) is difficult due to their
  complex structure and a high degree of occlusion. However, especially for the estimation of the position and size of fruits, it is
  necessary to avoid occlusions as much as possible and acquire sensor
  information of the relevant parts.  Global viewpoint planners exist
  that suggest a series of viewpoints to cover the regions of interest
  up to a certain degree, but they usually prioritize global coverage
  and do not emphasize the avoidance of local occlusions. On the other
  hand, there are approaches that aim at avoiding local occlusions,
  but they cannot be used in larger environments since they only reach
  a local maximum of coverage. In this paper, we therefore propose to combine a
  local, gradient-based method with global viewpoint planning to
  enable local occlusion avoidance while still being able to cover
  large areas.  Our simulated experiments with a robotic arm equipped
  with a camera array as well as an RGB-D camera show that this
  combination leads to a significantly increased coverage of the regions
  of interest compared to just applying global
  coverage planning.

\end{abstract} 

\section{Introduction}
\label{sec:intro}

Creating accurate 3D models of plants is difficult due to their
complex structure and lots of occlusions, e.g., caused by
leaves. Global viewpoint planning approaches typically evaluate the
information gain of multiple viewpoints to determine the next best
view. These methods can be used for large-scale viewpoint planning,
but since they consider the complete environment, they can fail in
complex environments with lots of local occlusions. On the other hand,
local viewpoint planning methods aim at avoiding local occlusions. For
example, Lehnert\etal\cite{lehnert_3d_2019} proposed to evaluate
differences in images from a camera array to find the
direction that maximizes the number of visible pixels of a given region of interest~(ROI) and therefore reduces occlusions. These methods can effectively avoid occlusions, but can only be
used locally. Once a local maximum is reached where the view on the
current target can no longer be improved, they cannot be applied to
plan further viewpoints.

In this paper, to get the best of both worlds we combine both global and local planning.
We use 3D Move to See~(3DMTS)~\cite{lehnert_3d_2019} to get a movement direction that improves the target visibility whenever occlusions of the ROI are detected. When 3DMTS reaches a
local maximum, we use global viewpoint planning 
\cite{zaenker2020viewpoint} to find the next best view and plan the
camera motion to this pose. The global viewpoint planning method builds up an
octree of the plants
with labeled ROIs, i.e., fruits, and uses this
octree to sample viewpoint candidates at frontier cells. To evaluate the viewpoints, the
framework applies a heuristic utility function that takes into
 account the expected information gain and automatically
 switches between ROI targeted sampling and exploration sampling,
to consider general frontier voxels, depending on the estimated
 utility. 

\begin{figure}
	%Compute heights
	\sbox\twosubbox{
		\resizebox{\dimexpr.93\linewidth}{!}{
			\includegraphics[height=3cm]{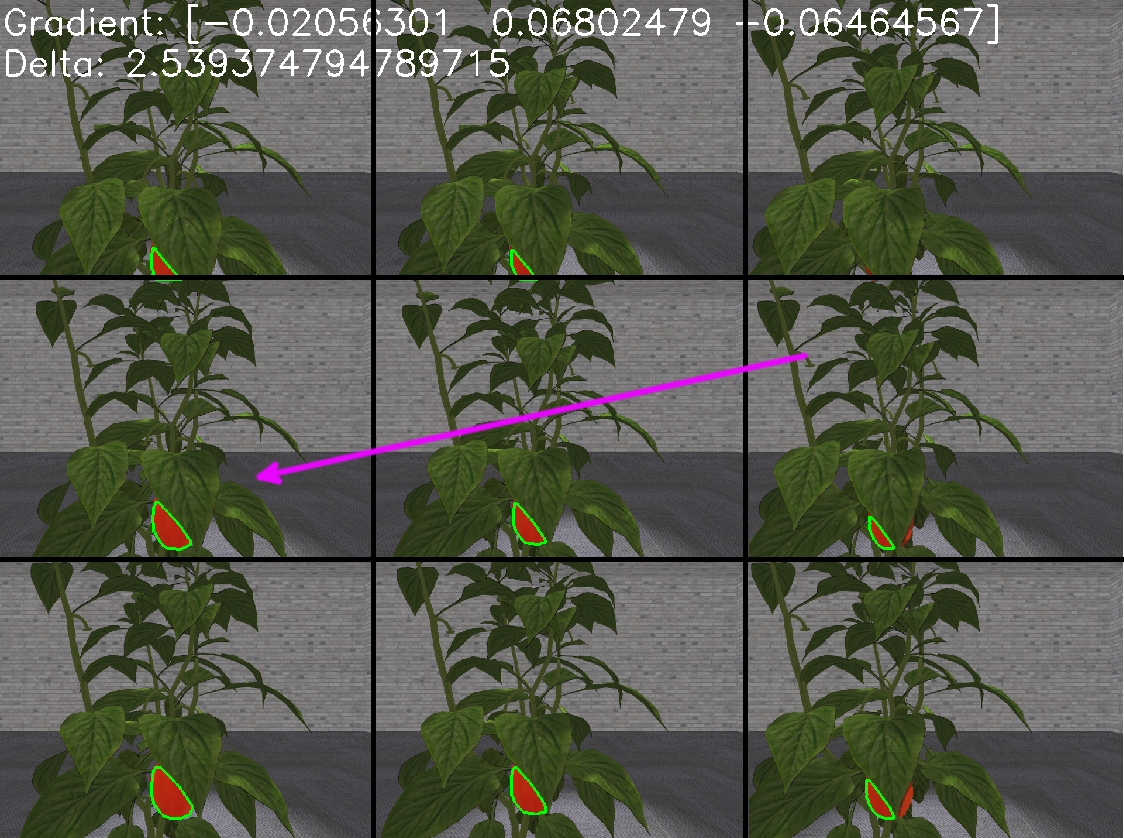}
			\includegraphics[height=3cm]{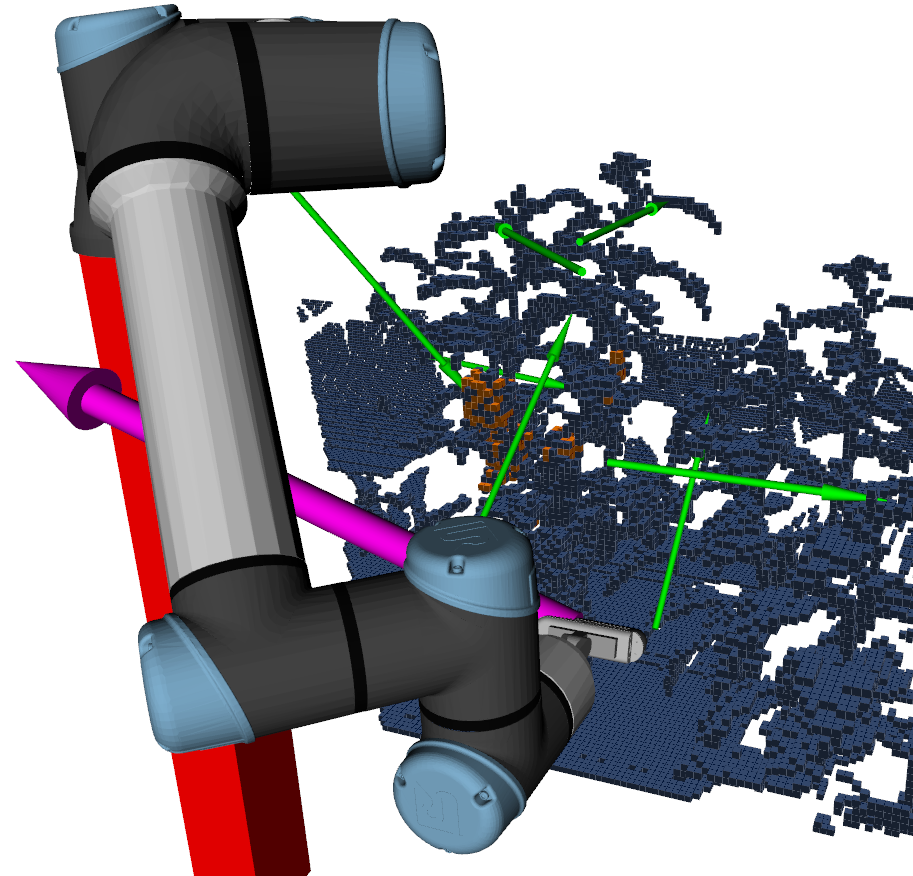}
		}
	}
	\setlength{\twosubht}{\ht\twosubbox}
	\centering
	\subcaptionbox{Result from Move to See\cite{lehnert_3d_2019}\label{fig:detection_example}}{
		\includegraphics[height=\twosubht]{images/m2s_combined}
	}\quad
	\subcaptionbox{Viewpoint selection\label{fig:planning_example}}{
		\includegraphics[height=\twosubht]{images/vpp_m2s_example}
	}
	
	\caption{Illustration of our combined approach. \textit{Left}:
          3DMTS~\cite{lehnert_3d_2019}
          detects a region of interest in all
          images of the camera array and suggests a moving direction
          for the
          camera to avoid occlusions. \textit{Right}: The camera
          mounted on a robotic arm either follows this gradient
          (pink), or selects one of the viewpoints from the global
          viewpoint planner~(green) depending on the expected improvement of the target visibility from following the gradient.
	}
	\label{fig:coverfig}
\end{figure}

\Cref{fig:coverfig} illustrates our proposed approach.
On the left image, 3DMTS uses a camera array to propose a gradient that improves the visibility of a partially occluded target.
On the right, our approach either decides to follow this gradient, if the expected information gain is high enough, or samples viewpoints with the global planner instead. The source code of our system is available on GitHub\footnote{\url{https://github.com/Eruvae/roi_viewpoint_planner}}.

Our contributions are the following: 
\begin{itemize}%[label=(\roman*)]
	\item Integration of 3DMTS~\cite{lehnert_3d_2019} with the
          global viewpoint planner~\cite{zaenker2020viewpoint}.
	\item Experimental evaluation in simulated scenarios of
          increasing complexity,
	comparing the new, combined approach with the
        previous global viewpoint planning method without 3DMTS in
        terms of number of detected ROIs as well as covered ROI volume.
\end{itemize} 

As the experiments with a robotic arm equipped with a camera
array as well as an RGB-D camera show, the presented combination leads to a
significantly increased coverage of the regions of
interest compared to just applying global viewpoint planning, since the
plants are covered in a locally more systematic way.

\section{Related Work}
\label{sec:related}

Viewpoint planning approaches can generally be divided into global
coverage path planners (CPP) that compute a complete viewpoint path
aiming at covering the desired area of a known environment, and local
next best view (NBV) planners that can be applied to unknown
environments. An example of the CPP paradigm is the method proposed
by \mbox{O{\ss}wald\etal\cite{oswald_efficient_2017}} that generates
viewpoints by casting rays from known object voxels towards free
space. The authors use a utility function that evaluates the viewpoint
candidates based on the number of visible object voxels and apply a
traveling salesman problem solver to compute the smallest tour of view
poses that cover all observable object voxels.  Similarly,
Jing\etal\cite{jing_coverage_2019} generate viewpoints based on the
maximum sensor range and compute viewing directions from the surface
normals of all target voxels within a certain range.  Afterward, the
authors proposed to randomly sample a set of points and connect nearby
points with a local planner to construct a graph.  Starting with the
current robot pose, the neighbors with the highest ratio of expected
information gain~(IG) and move cost are added to the solution path
until the desired coverage is reached.  There are a variety of use
cases for CPP methods, i.e., path planning for cleaning
robots~\cite{de1997complete} or covering a complete agricultural field with
machines for crop farming~\cite{oksanen2009coverage}.  Note that such
CPP approaches typically rely on a given representation of the
environment. However, in our agricultural use case for fruit size estimation, the environment
typically changes significantly with the growth of the plants and
their fruits.  Therefore, in our work, we do not assume a map of the
environment to be given beforehand. Instead, our framework builds a 3D
map of the plants during operation.

NBV approaches, in general, either rely only on current sensor
information or build a map of the environment while traversing it and
use this map to decide on the next view.  For example,
Lehnert\etal\cite{lehnert_3d_2019} use an array of cameras and
determine the size of a given target in each frame. Based on that,
their system computes a gradient to determine the direction for which
the visible area of the target is increased. In our work, we combine
this approach for local viewpoint planning to deal with occlusions
with global viewpoint planning using a map constructed during
operation~\cite{zaenker2020viewpoint}. In addition to the 3D~structure
of the plants, the map also represents the detected regions of
interest, i.e., fruits, which are used for sampling viewpoints aiming at
subsequently covering all fruits of a set of plants.
Wang\etal\cite{wang_autonomous_2019} use both current sensor
information and a built map for planning and proposed to combine an
entropy-based hand-crafted metric considering ray tracing from the
generated map with a local metric using a convolutional neural network
that takes the current depth image as input.  The two metrics are
combined to evaluate candidate poses generated in the vicinity of the
current camera position.
While this approach combines a local planning metric with a global, map-based algorithm, the learned image-based metric is not specifically targeted at avoiding occlusions.
Instead, it is meant to give an educated guess about the information gain in starting phases of planning, where information provided by the map is limited.

Monica\etal\cite{monica_humanoid_2019} consider the task of exploring
the environment around a given, single object of interest with a known pose
while performing 3D~shape reconstruction of the initially unknown
surrounding.  The authors also apply an exploration behavior for
unknown parts of the environment, but mainly to find new paths that may
enable observations of the object of interest.
Furthermore, Monica\etal\cite{monica2018contour} presented an NBV
method that samples viewpoints from frontiers to unknown space.
We use a similar technique to sample viewpoints for unexplored regions
independently of the detected ROIs when those have been sufficiently
explored.
\mbox{Palazzolo\etal\cite{palazzolo2018effective}} sample viewpoints
on the hull of the currently known map of an area of interest and
select the best point based on the estimated utility taking into
account the expected IG.  \mbox{Bircher\etal
  \cite{bircher2016receding}} proposed to use a rapidly exploring
random tree and estimate the exploration potential of a sequence of
points based on the
unmapped volume that can be covered at the nodes along the branches
of the tree.

Similar to the approach of Sukkar\etal\cite{sukkar_multi-robot_2019}
who detect apples as ROIs through color thresholding, we also rely on
automatic detection of the ROIs for viewpoint planning.
Sukkar\etal proposed to evaluate viewpoints based on a weighted sum of
exploration information, which is calculated from the number of
visible voxels that have not been previously explored, and ROI
information, which evaluates the visibility of ROIs from the selected
viewpoints.  This evaluation metric is then used to plan a sequence of
viewpoints for multiple robot arms by utilizing a decentralized Monte
Carlo tree search algorithm.

Some work also focuses specifically on viewpoint planning or analysis for fruit harvesting. Bulanon\etal\cite{bulanon2009fruit} compare the visibility of citrus fruits when recorded with different combinations of multiple fixed cameras.
Hemming\etal\cite{hemming2014fruit} place a single camera at multiple angles to determine which views provide the least occlusion for sweet peppers.
Both approaches deliver an offline analysis of the average visibility for different views in a practical application but do not provide a method to select viewpoints online based on an observation.
Kurtser and Edan \cite{kurtser2018use} propose an algorithm for sensing peppers that uses camera images to determine whether additional viewpoints are necessary and selects the next viewpoint from a predefined set with the best cost to benefit ratio.

In contrast to all the methods presented above, our approach is the
first one that combines an algorithm specifically targeted at avoiding local occlusions with a global, map-based approach that also considers the detected ROIs.

\section{System Overview}
\label{sec:approach}

\begin{figure*}[t] 	\centering 	\includegraphics[width=\linewidth]{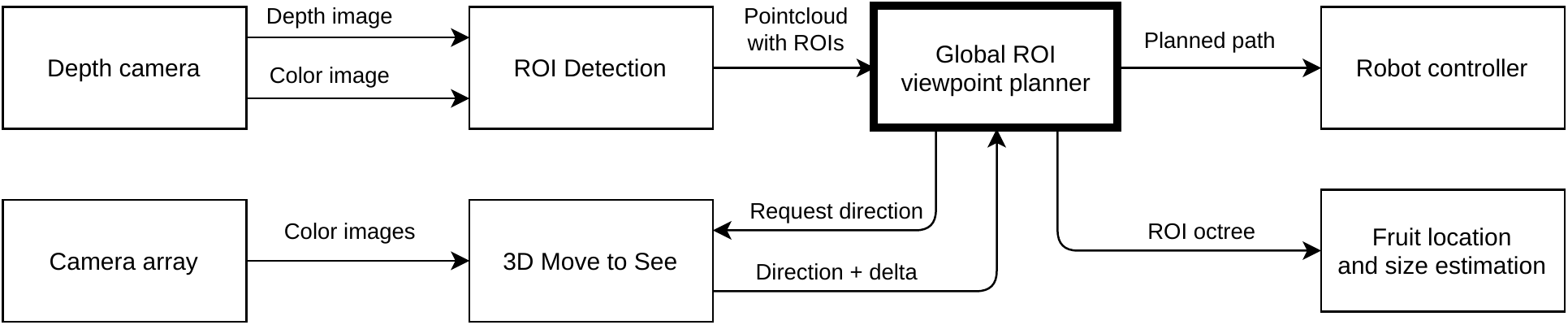} 	\caption{Overview of our system.
		See text for a detailed description.} 
	\label{fig:planner_overview}
\end{figure*} 

The goal of our work is to detect fruits of plants as regions of
interest (ROI) and acquire 3D data to estimate their volume. In
our application scenario, we currently use sweet pepper plants. In our
previously presented approach~\cite{zaenker2020viewpoint}, we used depth and color images of an RGB-D camera 
as an input to generate a 3D map with marked ROIs from which we sample
viewpoints. In the newly developed approach, we additionally apply 3D Move to
See~(3DMTS)~\cite{lehnert_3d_2019} to better deal with local
occlusions. 3DMTS relies on multiple camera images at
different positions as input.

\Cref{fig:planner_overview} shows an overview of our framework and the
integration of global viewpoint planning with 3DMTS. Global viewpoint
planning uses the depth and color image of the RGB-D~camera as
input, while the camera array outputs nine color images for 3DMTS. The
sweet peppers are detected in all color images~(see \secref{sec:approach_roidet}). For the global planner, the detection
is combined with the depth image to generate a point cloud with marked
ROIs. The generated point cloud is forwarded to the global planning module,
which uses the incoming information to build up a 3D map of the
environment in the form of an octree. The octree stores both, occupancy and
ROI information as described in~\secref{sec:approach_map}. The planner requests a
direction suggestion from 3DMTS, which is introduced
in~\secref{sec:move_to_see}. Depending on the change of the visibility
of the target in the different images of the camera array, the robot
arm with the cameras is moved
in that direction. Otherwise, targets and viewpoints are sampled and
evaluated using global viewpoint planning by sampling from frontier voxels, see
\secref{sec:approach_sampling} for details.

Using the MoveIt framework \cite{chitta2012moveit}, our system plans a
path for the robot arm to the best found viewpoint.
The planned path is then executed by the robot controller.

\section{Preliminaries}

\subsection{ROI Detection}
\label{sec:approach_roidet}

Both 3DMTS and the global viewpoint planning approach depend on
detected sweet peppers as the targets.
Since the plants in our simulation experiments only had red peppers,
we employ a simple color detection and find
red pixels using the HSI~(hue, saturation, color) color space.
For the global viewpoint planning, the detected regions in the color image are transformed to 3D by using the corresponding depth image.
A voxel grid filter is utilized to adjust the point clouds of the detected fruits to the resolution of the octree.
For 3DMTS, only the color images are used, the detections are grouped into clusters, and the largest cluster in the central image is used as the target.
For the other images in the camera array, the target is the cluster closest to the one in the central image.

\subsection{Octree for Viewpoint Planning}
\label{sec:approach_map}

For the global planner and fruit position and size estimation, our
system builds a custom octree to store 3D information about occupancy
and ROIs. The octree is updated after each movement step using a
point cloud with marked ROIs that are generated as described in
\secref{sec:approach_roidet}. The implementation is built on the
OctoMap framework \cite{hornung13auro} and the octree stores occupancy
and ROI log-odds.  The occupancy value is updated by casting rays from
the sensor origin to the points of the point cloud. For all nodes
encountered by the rays, the occupancy log-odds are decreased, while
for the nodes corresponding to the points of the point cloud, they are
increased.  Nodes with positive log-odds are considered as occupied,
nodes with negative log-odds as free.  The ROI probability is updated
only for the nodes directly on the scanned points.  For all points
marked as ROI, the log-odds are increased, while for all other points,
they are decreased.  All nodes with log-odds above a threshold are
considered as ROIs.

3DMTS does not use the octree itself, but the octree is updated with
new sensor information after each step taken by 3DMTS.

\section{Combined Viewpoint Planning}
\label{sec:approach_sampling}

We now describe our combined viewpoint planning approach.
In each step, we first attempt to avoid occlusions with 3DMTS if a target is visible (\secref{sec:move_to_see}).
If no target is visible or no direction that improves the view is
found, our system employs global planning instead and explores targets in the
vicinity of the ROIs (\secref{sec:sampling_roi}).
If no viewpoints are found that increase the information gain, more general exploration viewpoints to discover new ROIs are sampled (\secref{sec:sampling_explo}).
In both cases, we use a utility function to evaluate the viewpoints and determine the best one~(\secref{sec:approach_evaluation}).

\subsection{3D Move to See}
\label{sec:move_to_see}

\begin{figure}[t]
  \centering 	\includegraphics[width=\linewidth]{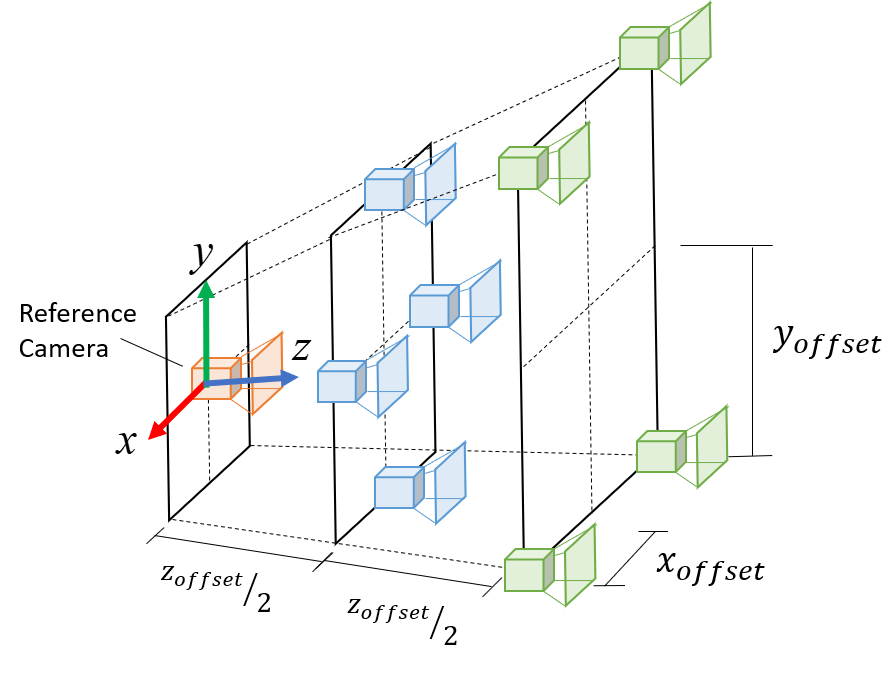} 	\caption{3DMTS
    camera array~(from  \cite{lehnert_3d_2019}). See text
    for description.} 
\label{fig:camera_layout_diagram}
\end{figure}

In order to enable avoiding occlusions locally, we integrated the
3DMTS approach of Lehnert\etal\cite{lehnert_3d_2019}, which uses a
camera array to find directions that locally improve the view on a target.
In each image of the camera array, fruits are detected as potential
targets, as discussed in \secref{sec:approach_roidet}. The cameras in the array are attached as illustrated in \figref{fig:camera_layout_diagram}, where the offset vector $\mathbf{o} = \left[0.027, 0.027, 0.03 \right]$ (for x, y and z respectively) is the same as in \cite{lehnert_3d_2019}.

We modified the original 3DMTS algorithm~\cite{lehnert_3d_2019} as
described in the following.
Originally, the objective function of 3DMTS consisted of a weighted sum of manipulability and target size.
In our application, manipulability is not relevant and so the objective function is simply the target size.

The target sizes are determined by computing the contours of connected fruit pixels. The contour area is then used as the target size.
Since it is possible that multiple targets are visible in an image,
the target with the largest contour area is selected in the reference image.
To find the corresponding fruits in the other camera array images, the target with the closest pixel coordinate center is selected, as long as its size is above a minimum threshold.
If no suitable target is found, the objective value for this frame is set to 0.

The computation of the gradient is carried out according to~\cite{lehnert_3d_2019}.
First, the constant directional vectors of the n (in our case 8) outer cameras are stored in $\matr{V}$
\begin{equation}
\matr{V_{(n\times3)}} = \begin{bmatrix} \bm{v}^0 & \hdots & \bm{v}^n\end{bmatrix}^T,
\end{equation}
the differences of the objective function values of the outer cameras to the reference image are computed as $\bm{\Delta{f}}$
\begin{equation}
\bm{\Delta{f}_{(n\times1)}} = \begin{bmatrix} \Delta{f_0} & \hdots & \Delta{f_n}\end{bmatrix}^T,
\end{equation}
and then the gradient is estimated as follows:
\begin{equation}\label{least_squares}
\bm{\nabla {f^*}_{(3\times1)}}=(\matr{V}^T\matr{V})^{-1}\matr{V}^T\bm{\Delta{f}}
\end{equation}
To estimate how much the view of the target improves by following the gradient, we added the computation of a scalar delta value to the original approach, which is used to decide when to switch to global planning. Each camera is assigned a weight depending on the computed gradient:

\begin{equation}
\matr{w_{(n\times1)}} = \matr{V}\cdot\bm{\nabla {f^*}}
\end{equation}
Each weight is the dot product of the corresponding camera vector with
the computed gradient. Therefore, cameras that correspond with the
direction of the gradient are weighted positively, and cameras in the
opposite direction are weighted negatively. The dot product of the weights and the objective function deltas is then computed to determine a scalar delta value $\Delta{f_{m}}$.
\begin{equation}
\Delta{f_{m}} = \matr{w}\cdot\bm{\Delta{f}}
\end{equation}
If 3DMTS is enabled, whenever the robot reaches a new viewpoint, the
planner requests the gradient $\bm{\nabla {f^*}}$ with the associated
weighted delta value $\Delta{f_{m}}$. If the latter is above a
threshold, the arm moves towards the gradient for a certain step
length. The process is repeated until delta is below the threshold,
the robot cannot move in the specified direction, or a maximum number
of steps is reached. After that, a new viewpoint is sampled using
global viewpoint planning~\cite{zaenker2020viewpoint}, which
is summarized in the following.

\subsection{ROI Targeted Sampling}
\label{sec:sampling_roi}

For the ROI targeted sampling, our system uses the frontiers of already detected ROIs as targets.
ROI frontiers are free voxels that have both an ROI and an unknown neighbor.
For each target, viewpoints are sampled by casting rays in a random direction with the specified sensor range as length.
If the resulting points are within the robot's workspace, they are
considered as potential next best view, and
their camera orientation is determined by rotating it so that the viewing direction aligns with the vector from the viewpoint to the target.
Furthermore, a ray is cast between the viewpoint and the target point.
If the ray passes an occupied node, the viewpoint is discarded, as the target is occluded.

\subsection{Exploration Sampling}
\label{sec:sampling_explo}

To discover new ROIs, the viewpoint planning considers general
frontier voxels within the so-called exploration sampling.  General
frontiers are determined as free cells with both occupied and unknown
neighbor cells. In this way, our system is able to find new ROIs after
all previous ROIs have been sufficiently explored.  After all targets
have been collected, potential viewpoints are sampled and their
directions are determined in the same way as for the ROI targeted
sampling.

\subsection{Viewpoint Evaluation}
\label{sec:approach_evaluation}

To evaluate the sampled views, we need to estimate their utility.
To do so, we first cast rays from the viewpoint within a specified
field of view of the sensor to estimate their information gain~(IG). Based on the rays, we calculate the IG for the encountered voxels using
the proximity count metric \mbox{\cite{zaenker2020viewpoint,delmerico18ijrr}}.
Each unknown voxel along the rays is given an initial weight of 0.5 and if it is within a specified distance~$d_{max}$ from a known ROI, the weight~$w$ is increased as follows
\begin{align}
w = 0.5 + 0.5 \cdot\frac{d_{max} - d}{d_{max}}
\end{align}
where~$d$ is its distance of the current voxel to the ROI. In this way, the weight of voxels close to observed ROIs is increased. Known voxels receive a weight of 0.

Considering the sum of the weights of the voxels along the rays $W_r$, the information gain of a viewpoint is defined~as
\begin{align}
	IG = \frac{1}{\vert R\vert}\sum_{r\in R}\frac{W_r}{N_r}
\end{align}
where ~$N_r$ is the total number of nodes on the considered ray.

In addition to the IG, we compute the cost $C$ for reaching the point.
Since computing the joint trajectory to reach the viewpoint is too
time-consuming to be done for each sampled view, we use the Euclidean distance of the camera to the point as an approximation.
Finally, the utility of a viewpoint is computed as the weighted sum of the IG and the cost scaled by a factor~$\alpha$: 
\begin{align}
U = IG - \alpha \cdot C
\end{align}

\begin{algorithm}[t]
	\SetAlgoLined
	m2sMoves = 0\;
	\While{True}{
		\If{m2sMoves $<$ maxMoves}
		{
			dir, delta = callMoveToSee()\;
			\If{delta $>$ deltaThresh}{
				moveInDir(dir)\;
				m2sMoves++\;
				continue\;
			}
		}
	    m2sMoves = 0\;
		chosenVps = sampleGlobalVps(nVps)\;
		\While{max(chosenVps) $>$ utilityThresh}
		{
			vp = extractMax(chosenVps)\;
			\If{moveToPose(vp)}
			{
				break\;
			}
		}
	    
	}
	\caption{Viewpoint planning}
	\label{algo:vpp}
\end{algorithm} 

\subsection{Viewpoint Selection}

In our proposed approach, we combine 3DMTS for local occlusion
avoidance with the two global sampling methods, ROI targeted sampling
and exploration sampling, as summarized in \algref{algo:vpp}. 
As long as 3DMTS does not exceed a maximum number of steps, our
planning system requests a direction suggestion from it.
If the returned delta value is above the specified threshold, the camera is moved in that direction.
Otherwise, global viewpoint sampling is executed using ROI targeted
and exploration sampling, and the best viewpoint based on the computed
utility is selected.
If no plan to reach this viewpoint with the arm can be computed, the next best viewpoint is used, until either the planner is successful, or no viewpoints above the utility threshold are left.
In the latter case, new viewpoints are sampled.

\section{Experiments}
\label{sec:exp}

\subsection{Simulated Environment}

We evaluated our viewpoint planning approach in simulated scenarios
with an RGB-D camera as well as a camera array mounted on a robotic
arm, i.e., the UR5e arm from Universal Robots.
The arm has six degrees of freedom and a reach of $85\,cm$.
To compute the workspace of the arm, the first 5 joints were sampled at a resolution of $10^{\circ}$.
The 6th joint was ignored, as it only rotates the camera and therefore does not change the viewpoint.
Like in \cite{zaenker2020viewpoint}, we set the resolution of the octree that stores occupancy and ROI information to 1 cm.

\begin{figure}
	%Compute heights
	\sbox\twosubbox{
		\resizebox{\dimexpr.93\linewidth}{!}{
			\includegraphics[height=3cm]{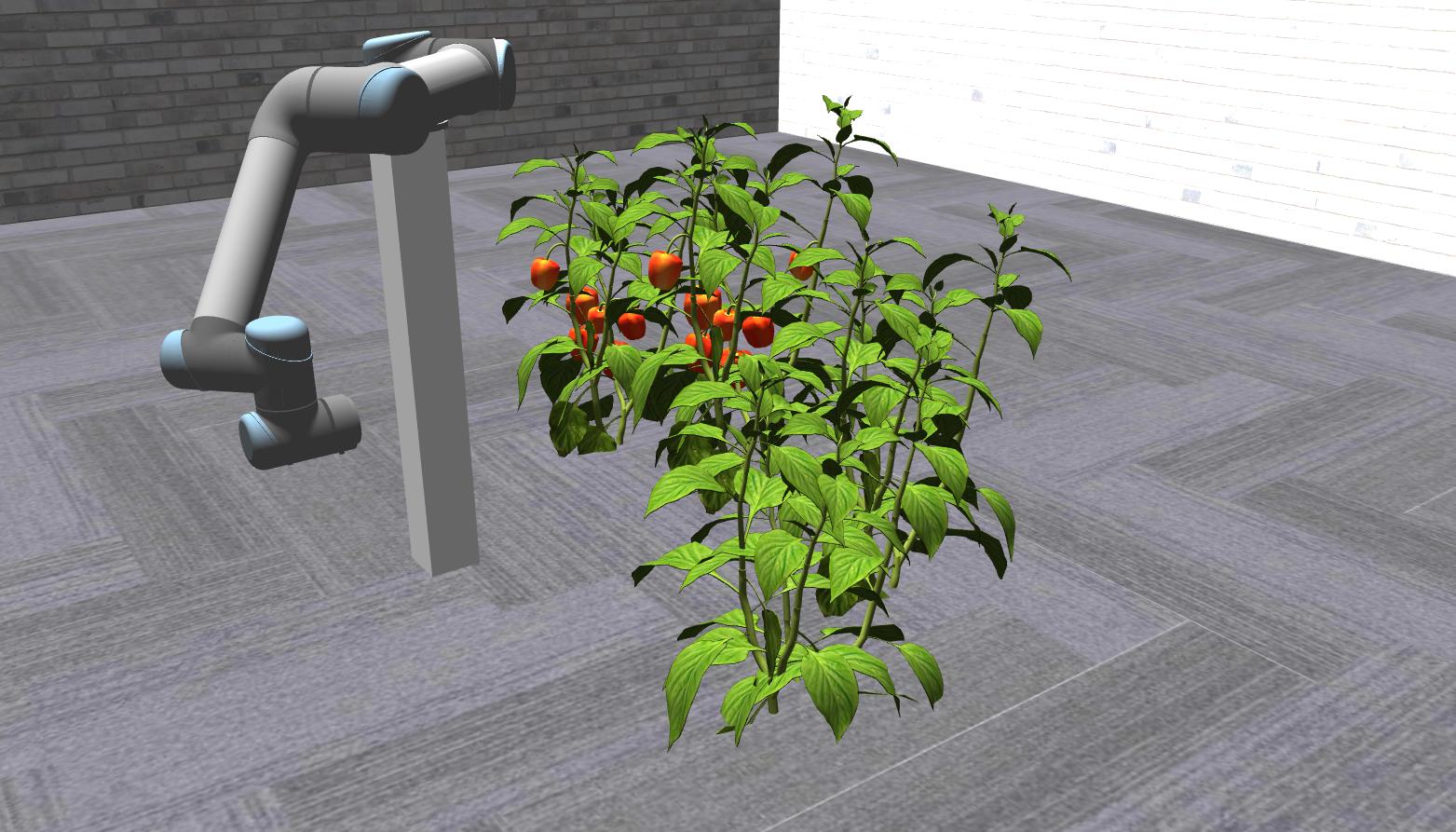}
			\includegraphics[height=3cm]{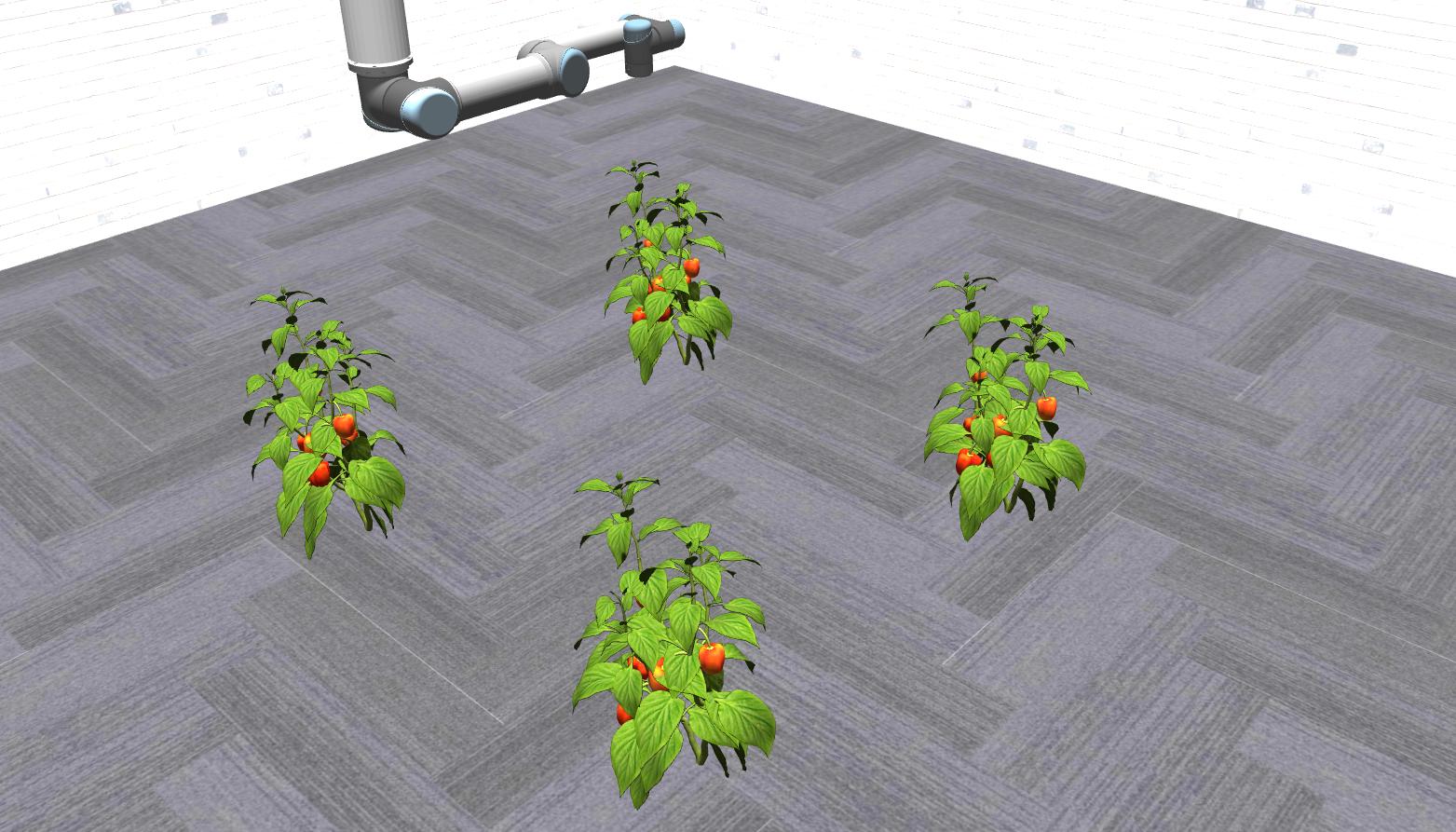}
		}
	}
	\setlength{\twosubht}{\ht\twosubbox}
	\centering
	\subcaptionbox{Scenario 1\label{fig:simulatedenv1}}{
		\includegraphics[height=\twosubht]{images/simulated_env_1}
	}\quad
	\subcaptionbox{Scenario 2\label{fig:simulatedenv2}}{
		\includegraphics[height=\twosubht]{images/simulated_env_2}
	}\quad\\[1ex]
    \subcaptionbox{Scenario 3\label{fig:simulatedenv3}}{
    \includegraphics[height=\twosubht]{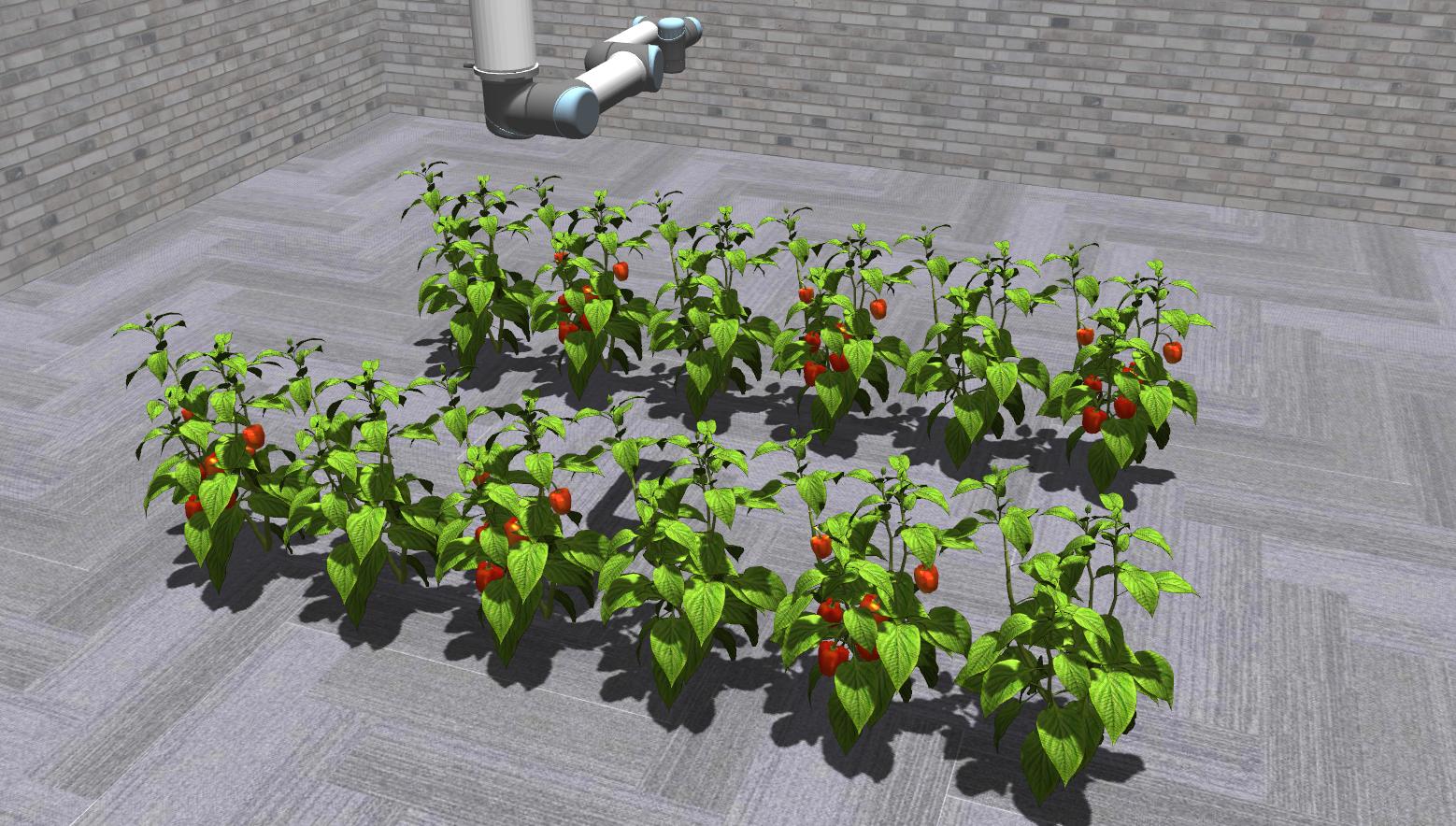}
    }
	
	\caption{Simulated environments. \textit{Left}: environment
		with four plants, two of which have seven fruits each and the other two do not have any fruits. The arm is placed on a static pole. \textit{Right}: Environment contains four plants with seven fruits each. The arm is hanging from
		the ceiling and can move within a $2\times2\,m$ square and extend up to $1.2\,m$ down.
		\textit{Bottom}: Same arm configuration as in Scenario~2, but the environment contains twelve plants in two rows, six of which have seven fruits each.
	}
	\label{fig:simulatedenvs}
\end{figure}

Three environments with different workspaces were designed for the simulated experiments.
The first two environments are the ones we used in \cite{zaenker2020viewpoint}, the third one was added as a more complex scenario.
In the first scenario, the arm is placed on top of a static $85\,cm$ high pole~(see \figref{fig:simulatedenv1}).
This allows the arm to exploit most of its workspace, except for the part blocked by the pole.
However, the movement possibilities are limited, as the arm cannot move itself.
Four plants were placed close to each other within reach of the arm.
To be able to explore a larger workspace, the arm was placed on a retractable, movable 3-DOF pole hanging from the ceiling for the second scenario~(see \figref{fig:simulatedenv2}). The pole can move within a $2\times2\,m$ square and extend up to $1.2\,m$ down.
With this setup, the arm is able to approach most of the potential poses in the simulated room.
Four plants were placed in the corners of the workspace, with a distance of 1.5 m between each.
The third scenario has the same setup for the arm, but instead of only four plants with a lot of space in between, two rows of six plants each are used. Only every second plant in each row has fruits.

\subsection{Evaluation}

For the evaluation, we use two of the  metrics from \cite{zaenker2020viewpoint}:
\begin{itemize} 
	\item \textit{Number of detected ROIs}: Number of found clusters 
	that can be matched with a ground truth cluster, which means that their center distance is smaller than~$20\,cm$.
	\item \textit{Covered ROI volume}: Percentage of the total volume of the ground truth that was detected, considering the 3D bounding boxes. 
\end{itemize}

Previously, we compared these metrics using the planning
time~\cite{zaenker2020viewpoint}. However, the planning time is dependent on several factors,
e.g., computation power and the number of tries until a successful
path is computed, which can lead to widely varying results, making a
comparison of the approaches difficult. Therefore, in our evaluation,
we use the plan length instead, which is proportional to the time it takes to execute the planned path.

For each trial, we let the planning system plan viewpoints until a
path duration of 120\,s was reached. We computed 20~trials for each approach in all three scenarios. To show the statistical significance, we performed a one-sided Mann-Whitney U test on the acquired samples.

\begin{figure} 	\centering
	\begin{subfigure}[b]{0.49\linewidth} 		\centering 		\includegraphics[width=\linewidth]{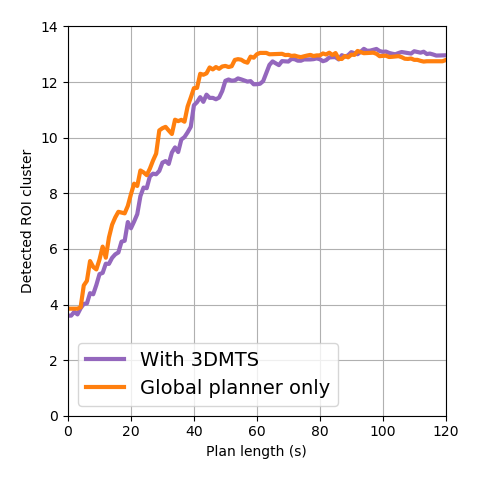} 		\caption{Detected ROIs} 
		\label{fig:res_w14_detroi}
	\end{subfigure} 
	\begin{subfigure}[b]{0.49\linewidth} 		\centering 		\includegraphics[width=\linewidth]{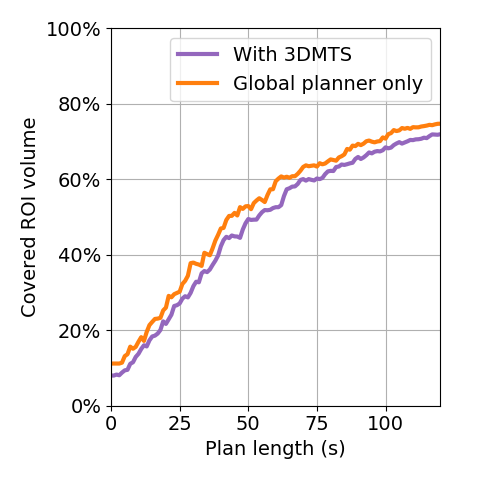} 		\caption{Covered ROI volume} 
		\label{fig:res_w14_covvol}
	\end{subfigure}	\hfill
	\caption{Results for Scenario 1.
		For each tested approach, 20 trials with a duration of 2 minutes each were performed.
		The plots show the average results.
		Here, the results with 3DMTS are slightly below the ones
                from the global planner, but the difference is not significant.} 
	\label{fig:res_w14}
\end{figure}

\begin{figure} 	\centering
	\begin{subfigure}[b]{0.49\linewidth} 		\centering 		\includegraphics[width=\linewidth]{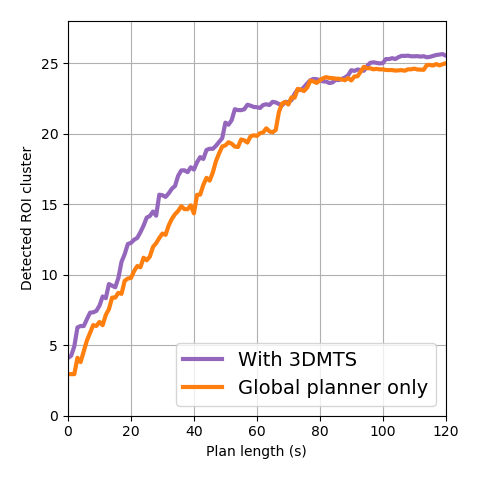} 		\caption{Detected ROIs} 
		\label{fig:res_w19_detroi}
	\end{subfigure}
	\begin{subfigure}[b]{0.49\linewidth} 		\centering 		\includegraphics[width=\linewidth]{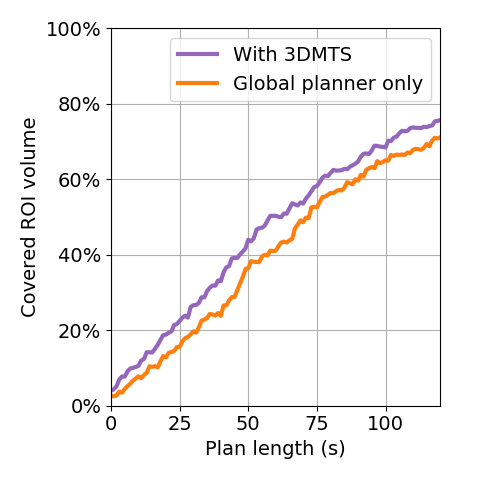} 		\caption{Covered ROI volume} 
		\label{fig:res_w19_covvol}
	\end{subfigure}	\hfill 
		 
	\caption{Results for Scenario 2.
		Like in Scenario 1, 20 trials were performed for each
                approach and the plots show the average
                results. As can be seen, the covered ROI volume of the
                combined approach is consistently above the one from the global
                planner. This difference at the end of the planning period is significant.}
	\label{fig:res_w19}
\end{figure}
\begin{figure} 	\centering
	\begin{subfigure}[b]{0.49\linewidth} 		\centering 		\includegraphics[width=\linewidth]{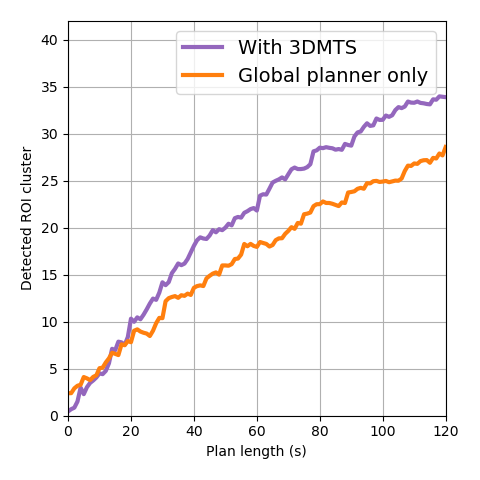} 		\caption{Detected ROIs} 
		\label{fig:res_w21_detroi}
	\end{subfigure}
	\begin{subfigure}[b]{0.49\linewidth} 		\centering 		\includegraphics[width=\linewidth]{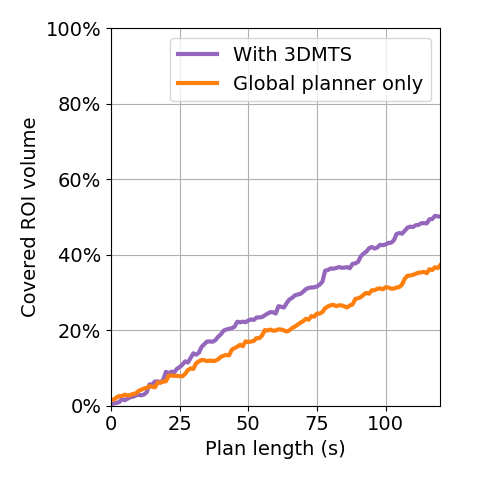} 		\caption{Covered ROI volume} 
		\label{fig:res_w21_covvol}
	\end{subfigure}	\hfill 
		 
	\caption{Results for Scenario 3. Here, the difference between
          the combined approach and the global planner is even
          clearer. The covered ROI volume is significantly higher and,
          furthermore, the number of detected ROIs is significantly  improved.}
	\label{fig:res_w21}
\end{figure}

\renewcommand{\arraystretch}{1.5}

\begin{table}
\centering 
\begin{tabularx}{\linewidth}{| l | l | X | X |} \cline{3-4}
\multicolumn{2}{c|}{} & With 3DMTS & Global pl. only\\ \cline{3-4}\hline
\multirow{2}{*}{\shortstack[c]{Sc. 1\\(14 ROIs)}}
& \# Det. ROIs & 13.0 $\pm$ 0.7 & 12.8 $\pm$ 0.7 \\ \cline{2-4}
& Cov. ROI vol. & 0.72 $\pm$ 0.06 & 0.75 $\pm$ 0.07 \\ \cline{1-4} 	
\multirow{2}{*}{\shortstack[c]{Sc. 2\\(28 ROIs)}}
& \# Det. ROIs & 25.6 $\pm$ 3.1 & 25.0 $\pm$ 3.3 \\ \cline{2-4}
& Cov. ROI vol. & \textbf{0.76 $\pm$ 0.14} & 0.71 $\pm$ 0.13 \\ \cline{1-4}
\multirow{2}{*}{\shortstack[c]{Sc. 3\\(42 ROIs)}}
& \# Det. ROIs & \textbf{33.9 $\pm$ 4.3} & 28.6 $\pm$ 5.4 \\ \cline{2-4}
& Cov. ROI vol. & \textbf{0.50 $\pm$ 0.13} & 0.37 $\pm$ 0.07 \\ \cline{1-4}
\end{tabularx}
\caption{Quantitative results over 20 trials. Bold values show a significant improvement compared to the other approach.
} 
\label{tab:res_table}
\end{table}

As can be seen from \figref{fig:res_w14} and \tabref{tab:res_table}, in Scenario 1 no significant improvement can be determined. The
approach with 3DMTS seems to perform slightly worse than the global
planner alone, but the difference is not significant. In Scenario
2~(see  \figref{fig:res_w19}), the approach with 3DMTS performs
slightly better than the global planner alone. The difference is
significant for the covered ROI volume according to the results in \tabref{tab:res_table}. Since the environment is larger, the additional local occlusion avoidance can cause the planner to stay with a single plant longer, while the global planner might quickly move on to the next plant, which promises a higher information gain. This can lead to a more efficient path using 3DMTS.

The difference is even more visible in Scenario 3, with more plants in
the environment. Here, 3DMTS leads not only to a significantly larger
covered ROI volume, but it also discovers significantly more fruits on
average~(see \tabref{tab:res_table}). Again, this can be explained with more efficient paths due to staying with the same plant for longer. Since a single plant has multiple fruits near each other, moving the camera around occluding leaves can lead to newly discovered fruits, while a global planner might already move on to other plants.

\section{Conclusions}
\label{sec:concl}

We introduced a novel viewpoint planning strategy that combines 3DMTS as an approach for local occlusion avoidance with a global planner to allow larger coverage.
Our experiments show that combining these strategies leads to an improved fruit coverage in large, complex environments compared to using only a global planner.

For future work, we plan to implement this approach on our existing robotic platform to enable more accurate scans of fruits in a commercial glasshouse environment.

\bibliographystyle{IEEEtran}
\bibliography{refs}

\end{document}